\documentclass[11pt,letterpaper]{article}
\usepackage{emnlp2017}
\usepackage{times}
\usepackage{latexsym}
\usepackage{url}
\emnlpfinalcopy

\def\emnlppaperid{***}

\title{ TextZoo, a New Benchmark for Reconsidering  Text Classification \Thanks{ $\dag$ means equal contribution. Need to be peer-reviewed}}

\author{Benyou Wang $^\dag$ \and Li Wang $^\dag$ \and Qikang Wei \and Lichun Liu\\
 Tencent \\ (wabywang,lilianwang,jaredwei,loringliu)@tencent.com
 }

\date{}

\begin{document}

\maketitle

\begin{abstract}
Text representation is a fundamental concern in Natural Language Processing, especially in text classification. Recently, many neural network approaches with delicate representation model (e.g. FASTTEXT, CNN, RNN and many hybrid models with attention mechanisms) claimed that they achieved state-of-art in specific text classification datasets. However, it lacks an unified benchmark to compare these models and reveals the  advantage of each sub-components for various settings. We re-implement more than 20 popular text representation models for  classification in more than 10 datasets. In this paper, we reconsider the text  classification task in the perspective of neural network and get serval effects with  analysis of the above results.

\end{abstract}

\section{Introduction}

In Natural Language Processing or text related community, effective representation of textual sequences is the fundamental topic for the up-stream tasks. Traditionally,  bag-of-word models (TFIDF or language model) with vocabulary-aware vector space tends to be the main-stream approach, especially in the task with long text (e.g. ad hoc retrieval with long document, text classification for long sentence). However, it tends to get pool performance in the tasks with short-text sentence (text classification for relatively short sentence, Question answering, machine comprehension and dialogue system), which there are little word-level overlaps in bag-of-word vector space. Distributed representation \cite{le2014distributed}  in a fixed low-dimensional space trained from large-scale corpus have been proposed to enhance the features of text, then break through the performance bottleneck of bag-of-words models in short-text tasks.
With combination of Conventional Neural Network (CNN) \cite{kalchbrenner2014convolutional}, Recurrent Neural Network (RNN), Recursive Neural Network \cite{socher2013recursive} and Attention, hundreds of models had been proposed to model text for further classification, matching \cite{fan2017matchzoo} or other tasks.

However, these models are tested in different settings with various datasets, preprocessing and even evaluation. Since subtle differences may lead to large divergence in final performance. It is essential to get a robust comparison and tested in rigid significance test. Moreover, models with both effective and efficient performance is impossible due to the No-Free-Lunch principle.  Thus  each  model should be considered in a trade off between its effectiveness and efficiency.

Out contribution is
\begin{enumerate}
\item A new open-source benchmark of text classification  \footnote{ Code in https://github.com/wabyking/TextClassificationBenchmark}  with more than 20 models and 10 datasets.
\item Systemic reconsideration of text classification in a trade off.
\end{enumerate}

\section{Models}

Models are shown as follow:

\noindent{\bf Fastext\cite{joulin2016bag}.} Sum with all the input embedding.

\noindent{\bf LSTM.}                        Basic LSTM \cite{hochreiter1997long} over the input embedding sequence.

\noindent{\bf BiLSTM.}                      LSTM with forward and backward direction.

\noindent{\bf StackLSTM.  }                 LSTM with multi layers.

\noindent{\bf Basic CNN.}                   Convolution over the input embedding \cite{kalchbrenner2014convolutional}.

\noindent{\bf Multi-window CNN \cite{severyn2015learning}.} Padding input embedding for a fixed size and concat the feature maps after convolution.

\noindent{\bf Multi-layer CNN.}             CNN with multi layers for high-level modelling.

\noindent{\bf CNN with Inception.}          CNN with Inception mechanism \cite{szegedy2015going}.

\noindent{\bf Capsules.}                    CNN with Capsules Networks \cite{NIPS2017hinton} .

\noindent{\bf CNN inspired by Quantum.}     Neural representation inspired by Quantum Theory \cite{zhang2018end,niu2017bi}.

\noindent{\bf RCNN \cite{lai2015recurrent}.} LSTM with pooling mechanism.

\noindent{\bf CRNN \cite{zhou2015c}.}       CNN After LSTM .

\section{Dataset}

There are many datasets as showed in Tab. \ref{tab:dataset}
\begin{table}
\small
\centering
\begin{tabular}{|l|c|c|c|c|}
\hline \bf Dataset & \bf Label  & \bf Vocab. & \bf Train  & \bf Test \\ \hline
20Newsgroups \footnote{ in qwone.com/jason/20Newsgroups/ } &   &  && \\
SST \footnote{4nlp.stanford.edu/sentiment/} & & && \\
Trec   &     & &      &\\
IMDB   &  2  & &25000 & 25000\\
SST-1  &  5  & &8544  &2210 \\
SST-2  &  2  & &6920  &1821\\
SUBJ   &  2  & &9000  &1000\\

\hline
\end{tabular}
\caption{\label{tab:dataset} Font guide.}
\end{table}

\subsection{Evalution}
We adopt the Precision as the final evaluation metrics, which is widely used in the  classification task.

\section{Conclusion}

As claimed in the introduction, A benchmark for text classification have been proposed to systemically compare these state-of-art models.  Performance, Significance test, Effectiveness-efficiency Discussion, Case study, comparison between RNN and CNN, Embedding sensitive needs to be done.

\section*{Acknowledgments}

\bibliography{emnlp2017}
\bibliographystyle{emnlp_natbib}

\end{document}